\pdfoutput=1

\documentclass{article}
\usepackage{arxiv}

\usepackage[utf8]{inputenc} 
\usepackage[T1]{fontenc}    
\usepackage{hyperref}       
\usepackage{url}            
\usepackage{booktabs}       
\usepackage{amsfonts}       
\usepackage{nicefrac}       
\usepackage{microtype}      
\usepackage{lipsum}
\usepackage{graphicx}
\usepackage{comment}
\graphicspath{ {./images/} }
\usepackage{multirow}
\newcommand{\argmin}{\mathop{\mathrm{argmin}}}

\title{Few-Shot Classification of Skin Lesions from Dermoscopic Images by Meta-Learning Representative Embeddings }

\author{
 Karthik Desingu \\
  Dept. of Computer Science and Engineering\\
  Sri Sivasubramaniya Nadar College of Engineering\\
  Chennai, TN, India \\
  \texttt{karthik19047@cse.ssn.edu.in} \\
   \And
 Mirunalini P. \\
  Dept. of Computer Science and Engineering\\
  Sri Sivasubramaniya Nadar College of Engineering\\
  Chennai, TN, India \\
  \texttt{miruna@ssn.edu.in} \\
  \And
 Aravindan Chandrabose \\
  Dept. of Information Technology\\
  Sri Sivasubramaniya Nadar College of Engineering\\
  Chennai, TN, India \\
  \texttt{aravindanc@ssn.edu.in} \\
}

\begin{document}
\date{}
\maketitle

\begin{abstract}
Annotated images and ground truth for the diagnosis of rare and novel diseases are scarce. This is expected to prevail, considering the small number of affected patient population and limited clinical expertise to annotate images. Further, the frequently occurring long-tailed class distributions in skin lesion and other disease classification datasets cause conventional training approaches to lead to poor generalization due to biased class priors. Few-shot learning, and meta-learning in general, aim to overcome these issues by aiming to perform well in low data regimes. This paper focuses on improving meta-learning for the classification of dermoscopic images. Specifically, we propose a baseline supervised method on the meta-training set that allows a network to learn highly representative and generalizable feature embeddings for images, that are readily transferable to new few-shot learning tasks. We follow some of the previous work in literature that posit that a representative feature embedding can be more effective than complex meta-learning algorithms. We empirically prove the efficacy of the proposed meta-training method on dermoscopic images for learning embeddings, and show that even simple linear classifiers trained atop these representations suffice to outperform some of the usual meta-learning methods.

\keywords{Dermoscopic Images, Few-Shot Learning, Meta Learning, Embedding Network} 
\end{abstract}

\section{Introduction}
Annotated data is essential for any supervised learning algorithm, and the availability of large amounts of annotated data helps deep learning methods achieve quantum leaps in classification tasks. Deep learning networks requires large amount of data to be robust enough to generalize feature learning for classification during training. Even though large amount of data is available in the digital world, availing annotated data is very difficult. In the medical domain especially, obtaining annotated data for rare or newly emerging diseases is even harder: employing medical experts to annotate the data, and manually labeling the large amount is data is time consuming and expensive. Furthermore, in some applications, the data across the classes is not distributed evenly. This leads to biased learning. Hence, in this research work, we propose a meta-learning approach that helps in rapid model adaptation to deal with data scarcity and unbalanced class distribution of datasets. 

Deep learning models have proven their value in the various aspects of the automation of medical image analysis; skin cancer type classification from dermatological images is one such area. Automated classification of skin cancer type is complex, owing to the similarity between the different types of lesions; the limited availability of annotated data for some cancer types due to the relative rarity of some of these cancer types; and the characteristic uneven distribution of samples across different classes due to similar reasons. When deep learning models are applied on such data, they were often biased towards the classes with large number of samples, and failed to generalize well for skewed classes. Clearly, there is a need to facilitate models to learn from small amounts of annotated data, and readily adapt to new classes with small sample sizes.

In this paper, we present a baseline learning approach for few-shot image classification in the meta-learning context. This approach has been underappreciated in the literature thus far. To the best of the authors' knowledge, this is the first application of this method for skin image classification in the few-shot regime. This approach consistently outperforms the protoypical network on two different datasets, even with the same backbone network. The method has been applied using multiple backbone networks --- simple convolutional neural network, ResNet18 and ResNet50 \cite{resnet}. Feature embeddings learnt through this method are, in general, more representative and are able to generalize well to unseen classes in few-shot tasks.

\section{Related Work}

There is limited availability of labeled data. Even annotated data is often characterized by uneven distribution of samples across classes. Although deep learning models march towards great heights in several challenging tasks \cite{szegedy,he}, it demands large amounts of data \cite{deng,Zhou} to achieve such performance. Meta-learning algorithms, however, can learn from limited data and produce comparable prediction performance. This not only solves the annotation insufficiency issue, but also works well on unbalanced datasets, learning to represent the minority classes well. Several researchers have developed, experimented with, and implemented meta-learning successfully to obtain high performance with limited data. Few-Shot Learning (FSL), a form of meta-learning targets the problem of data scarcity by learning a model using set of base classes, and adopts to new classes with just a few samples \cite{finn,p1,p2}. FSL uses a metric learning approach to achieve this \cite{afham,andrei}.

A model-agnostic meta-learning method was proposed in \cite{finn} which quickly adapt to new tasks. Metric learning based on deep neural features was proposed in \cite{p1} that maps a small labeled support set and an unlabeled example to its label that helps to adapt to new novel classes. A prototypical network was formulated in \cite{p2} that uses the Euclidean distance computed for a prototype representation of each class. Furthermore, a transformer based bi-directional decoding mechanism was used to learn correlations between same class samples in order to connect visual clues with semantic descriptions was proposed in \cite{afham}. Latent Embedding Optimization (LEO) was proposed to learn low-dimensional model parameters and performs meta-learning in this space \cite{andrei}. One-shot learning setting made correct predictions using Siamese neural networks which employ a unique structure to rank the inputs based on the similarities \cite{Koch}. Matching networks applied an LSTM based context encoder to match query and support set images \cite{oriol} in one-shot learning paradigms. An LSTM based meta-learner model was proposed to learn perfect a optimization algorithm, that helps to train another neural network classifier in the few-shot regime \cite{ravi}. An edge-labeling graph neural network (EGNN) was proposed to predict the edge-labels, which in turn enables explicit clustering by iteratively updating the edge-labels by exploitation of intra-cluster similarity and inter-cluster dissimilarity \cite{kim}.


\section{Method}

\label{sec: methods}

This section sets up and defines the preliminaries about the few-shot problem, and the meta-learning problem in general. The specific training-testing approach adapted is detailed in subsequent sections. To facilitate easy comparison, the same notations used in \cite{tian} for formulations are used in this paper.

\subsection{Meta-Learning: Problem Formulation}
\label{sec:meta-intro}
The learning problem is presented as a meta-learning task with two phases --- the meta-training and the meta-testing phase. The meta-training phase is based on the meta-training set $T$. $T$ is defined by,
$T = {(D^{train}_i, D^{test}_i )}^I_{i=1}$, where each tuple $(D^{train}_i, D^{test}_i)$ represents a training and a testing dataset for a single episode of the task. Each dataset contains a few examples. The training examples $D^{train} = {(x_t, y_t)}^T_{t=1}$ constitute the support set $S$, and the testing examples $D^{test}$ = ${(x_q , y_q)}^Q_{q=1}$  form the query set $Q$ of the episode. Both $D^{train}$ and $D^{test}$ are sampled from the same distribution.

A base learner $A$, defined by $y_{*} = f_\theta(x_{*})$, is trained on $D^{train}$ and used as a predictor on $D^{test}$. For image classification problems, $x_{*}$ is characterized by very high dimensionality, consequently a very high variance. Hence, $D^{train}$ and $D^{test}$ are first projected onto a feature space using a feature embedding model that can be formulated as $\phi_{*} = f_\phi(x_{*})$. 
The key intent of the meta-training phase is to learn a good embedding model, so that the base learning performs with minimal loss on a set of query tasks ($D^{test}$). This can be stated formally as,
\begin{equation}
\phi = \argmin_\phi E_T [L^{meta}(D^{test}; \theta, \phi)]
\end{equation}
where $\theta = A(D^{train}; \phi)$ is the base learner, and is fixed through training.

The embedding model $\phi$ is fixed when training the base learner in each episode. The objective of the base learner on each task can then be modeled as,
\begin{equation}
\theta = A(D^{train}; \phi) \\
    = \argmin_\theta L^{base}(D^{train}; \theta, \phi) + R(\theta)
\end{equation}
where R is the regularization term, and L represents the loss function for training the base learner.

A trained meta-learning model is then evaluated on a set of held-out tasks termed as the meta-testing set, $S = {(D^{train}_j, D^{test}_j )}^J_{j=1}$. Once again, $D^{train}$ and $D^{test}$ form the support and query sets for the meta-testing episode, respectively. The evaluation is done over a distribution of the test tasks, represented by,

\begin{equation}
\theta = E_S[L^{meta}(D^{test}; \theta, \phi)]
\end{equation}

 where $\theta = A(D^{train}; \phi)$.

\subsection{Prototypical Network Approach}
To prepare a baseline for comparison with the proposed meta-learning approach, the prototypical networks are adopted in this work. Prototypical network is a distance metric based meta-learning technique which computes the mean vector to represent each class of a support set. This vector is termed the prototype for the particular class in the support set. 

Conforming to the aforementioned formulation, a subset of classes is randomly selected to compose one training episode. For each training episode, a support set S, and a query set Q are sampled. For a N-way K-shot setting, the support set is sampled to contain N different classes, with K examples per class. Prototypical networks first use the embedding function $f_\phi$ to map the images to a common feature space. Each class c in S is then represented in the embedding space by a prototype vector $V_c$, which is computed as the mean vector of the embedded inputs for all the samples in S, corresponding to class c, as follows:

\begin{equation}
V_c = \frac{1}{K} \sum_{(x_j, y_j) \in S_c} f_\phi(x_{j})
\end{equation}

Each sample in the query set is then mapped to the embedding space. The Euclidean distance (equation \ref{eq:euclidean_dist} — Minkowski distance of second order — is computed between the prototype vector $V_c$ of each class in S, and the query example. The query example is classified under the class that it is closest in distance to. The embedding network is trained by back propagating the negative log loss of the query example’s distance from its classifier class vector. 

\begin{equation}
    \label{eq:euclidean_dist}
    d(p, q) = \sqrt{\sum_{k=1}^K (p_k - q_k)^2}
\end{equation}
where, $p$ and $q$ denote the prototype and query images' embedding vectors, $K$ denotes the dimensionality of the embedding vectors, and $d$ denotes the Euclidean distance.

During the meta-testing phase, the embedding network is fixed, with the rest of the method for classifying a query example remaining the same --- map the support and query sample to the embedding space, compute prototype vectors for each class in the support set, find the euclidean distance of the query sample from each prototype vector and classify under the class with least distance. 

\subsection{Pre-train and Base Learner Approach}
As presented in Section \ref{sec:meta-intro}, the meta-training phase intends to learn an embedding model $f_\phi$, that projects the input on to a lower dimensional, but a representative, feature space. At the crux of the work presented in this paper, lies the importance of the generalizability and characteristic nature of this embedding model. It should be readily transferable to any new task, such that a projection onto the learned feature space solves the new set of tasks with very limited examples.

A great number of meta-learning algorithms aim to learn the embedding model by training them on training tasks, modeled very similarly to the target tasks — episodic learning used in prototypical networks, for instance \cite{laenen}. However, this paper follows the approach in \cite{tian}, and posits that a model that is pre-trained on a classification task, learns to produce highly representative and discriminative embeddings that can be leveraged by the base learner effectively, even when transferred to a new task. 

\begin{figure}[!htb]
    \centering
    \includegraphics[width=0.95\textwidth]{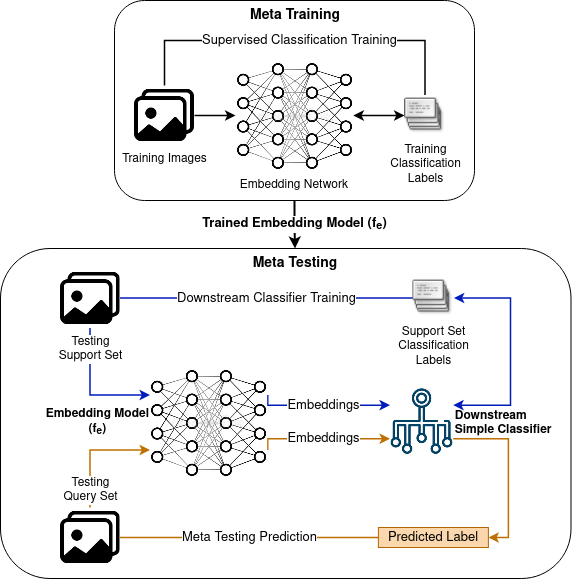}
    \caption{Flow diagram depicting the proposed meta training and testing approach using an embedding network.}
    \label{fig:metalearning-flow}
\end{figure}

To this end, the training datasets from each episode of the meta-training phase are merged into one large dataset $D^{new}$, given as,

\begin{equation}
D^{new} = {(x_t, y_t)}^K_{k=1} \\
 = \bigcup {D^{train}_1, ..., D^{train}_i, ..., D^{train}_I}
\end{equation}

where $D^{train}_i$ is the training dataset from T for episode i. 

The embedding model then becomes,

\begin{equation}
\phi = \argmin_\phi L^{ce}(D_{new}; \phi)
\end{equation}
where $L^{ce}$ denotes the cross-entropy loss between predictions and ground-truth labels. In effect, this reduces to the task of training a simple supervised image classifier. 

The meta-testing phase, however, is modeled quite similarly to most meta-learning setups. Concretely, for every task ($D^{train}_j$ , $D^{test}_j$) that is sampled from meta-testing distribution, we train a base learner on $D^{train}$j that makes predictions on $D^{test}_j$. In this paper, the base learner is modeled as a linear model The linear model is attributed with parameters $\theta = {W , b}$ to include a weight term W and a bias term b, that are learned through,

\begin{equation}
\theta = \argmin_{W,b} L^{ce}_t(Wf_\phi(x_t) + b, y_t) + R(W, b)
\end{equation}
where R is the regularization term.

In addition, this paper also presents alternative base learners that leverage the learned embedding model --- decision trees, and nearest-neighbor classifiers with the Minkowski distance metric of the second order.

Concretely, the embedding model is learned during the meta-training phase by back propagating the cross-entropy loss of a classification task, formulated using the complete set of training examples from each episode --- namely $D^{new}$. The learned embedding network is the fixed, and a base learner --- linear, tree, or any other feature classifier — is learned for each episode of the meta-testing phase. Particularly, the supervised learner trains with the few examples available in the support set $D^{train}$ of an episode in S, and makes predictions for the corresponding query set $D^{test}$ of the episode. Hence, a powerful embedding model would allow the base learner to learn to classify on a new task, with just a few examples. Hence, the significant difference lies in the approach adopted to learn the embedding model during meta-training, while meta-testing still adopts episodic learning, but while training a downstream base classifier for each episode. The proposed approach is similar to the methods adopted in \cite{andrei,boris} and differs from \cite{guneet,anirudh} which additionally fine-tune the embedding model $f_\phi$ during the meta-testing phase. The proposed learning approach is depicted in Figure \ref{fig:metalearning-flow}.

\section{Experiments}

This paper conducts experiments on two widely accepted benchmark datasets for skin lesion classification: the ISIC 2018 Task 3 Dataset  and \cite{isic18t3} the Derm7pt Dataset \cite{derm7pt}. All experiments in this research can be reproduced through the implementation that will be made available at \url{https://github.com/karthik-d/Few-Shot-Learning-Skin-Analysis}.

\subsection{Datasets}

\begin{figure}
    \centering
    \includegraphics[width=0.8\textwidth]{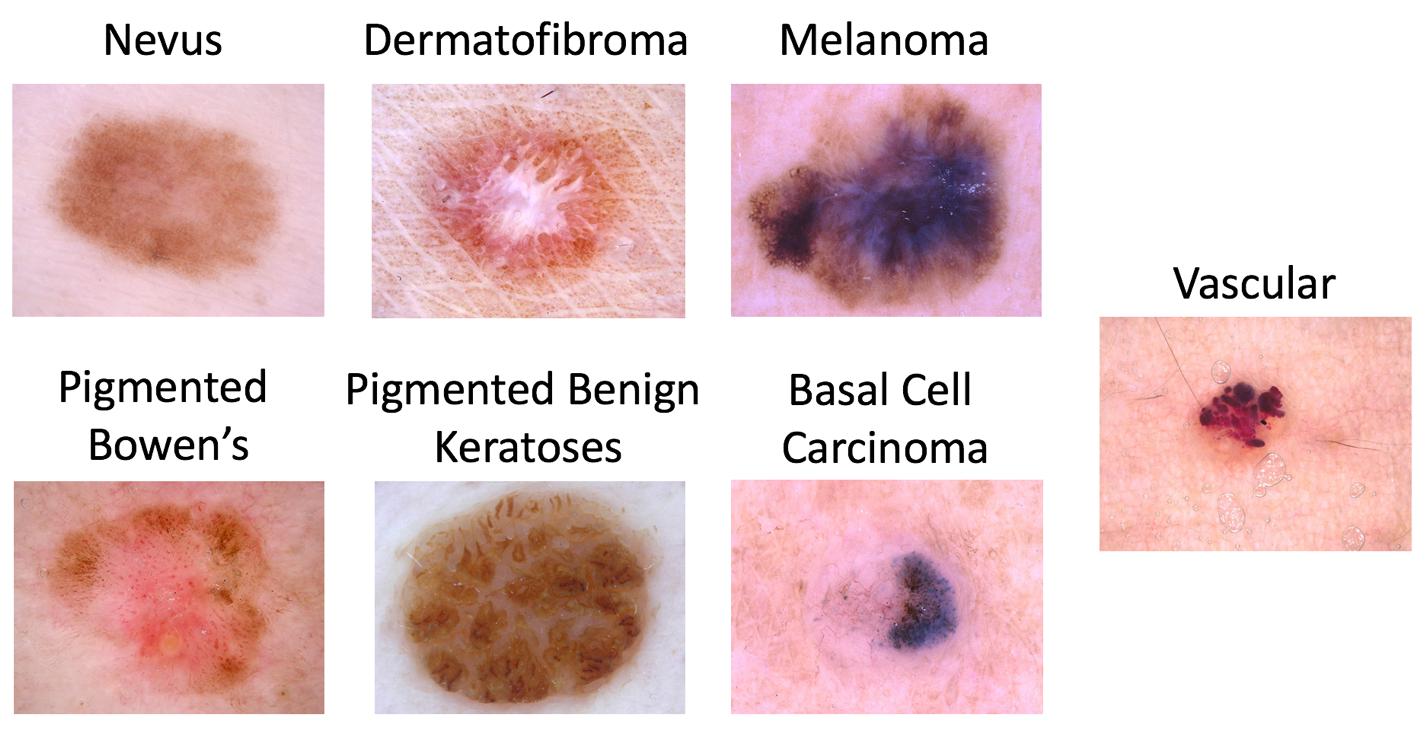}
    \caption{Samples images from each class of the \textbf{ISIC-2018 dataset}.}
    \label{fig:2}
\end{figure}

\textbf{ISIC-2018 Task-3} \cite{isic18t3}: This dataset consists of 10,015 dermoscopic images. These are labeled by expert pathologists into one of the seven skin lesion categories. Based on a standard train-test split of 80-20, a total of 7,515 images compose the train set, while the remaining 2,500 form the test set. For the purpose of experiments, the authors resized the images from $600 \times 450$ pixels to $ 224 \times 224$ pixels, and chose four and three classes in the meta-train and meta-test sets respectively, to formulate the meta-learning classification episodes. Figure \ref{fig:2} shows sample images from the dataset. Table \ref{table:isic_split} represents the data split.

\begin{table}
\begin{center}
\caption{Split of the classes in the \textbf{ISIC dataset} into training and testing sets.}
\label{table:isic_split}
\begin{tabular}{llll}
\hline\noalign{\smallskip}
\textbf{Class Abbr}. & \textbf{Class Name} & \textbf{Number of Images} & \textbf{Data Split} \\
\hline
\noalign{\smallskip}
NV & Nevus & 6741 & Train \\
MEL & Melanoma & 1119 & Train \\
BKL & Keratosis & 1101 & Train \\
BCC & Basal Cell Carcinoma & 517 & Train \\
AKIEC & Actinic & 331 & Test \\
VASC & Vascular & 143 & Test \\
DF & Dermatofibroma & 116 & Test \\
\hline
\end{tabular}
\end{center}
\end{table}

\begin{figure}
    \centering
    \includegraphics [width=0.6 \textwidth]{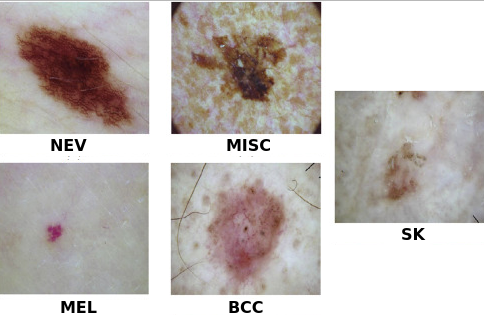}
    \caption{Sample images from each class of the \textbf{Derm7pt dataset}.}
    \label{fig:1}
\end{figure}

\textbf{Derm7pt} \cite{derm7pt}: This dataset that includes over 2000 clinical and dermoscopy color images belonging to 20 distinct diagnostic classes, further grouped into 5 super classes. It includes structured metadata to benchmark the training and evaluation of automated diagnosis systems. The dataset provides a 7-point skin lesion malignancy checklist, and bases the predictions on this. The original image size is $768 \times 512$ pixels. Once again, for experimentation, the dermoscopic images are used, and these are resized to $224 \times 224$. The standard train-test splits benchmarked by the dataset providers is used in these experiments. To facilitate evaluation of the proposed meta-learning algorithm, the 5 super classes are considered. The 3 classes with the least number of images per class are used for the meta-training phase, and the other 2 are used for meta-training.  Table \ref{table:derm7pt_split} represents the data split and the sample image from Derm7pt is represented in the \ref{fig:1}.

\begin{table}
\begin{center}
\caption{Split of the classes in the \textbf{Derm7pt dataset} into training and testing sets.}
\label{table:derm7pt_split}
\begin{tabular}{llll}
\hline\noalign{\smallskip}
\textbf{Class Abbr}. & \textbf{Class Name} & \textbf{Number of Images} & \textbf{Data Split} \\
\hline
\noalign{\smallskip}
NEV & Nevus & 575 & Train \\
MEL & Melanoma & 252 & Train \\
MISC & Miscellaneous & 97 & Test \\
SK & Seborrheic Keratosis & 45 & Test \\
BCC & Basal Cell Carcinoma & 42 & Test \\
\hline
\end{tabular}
\end{center}
\end{table}

Both the datasets are characterized by a heavy class imbalance. The classes with the least number of data samples are moved to the test set and used as the domain to sample the meta-test tasks, to model the real-world setting and assess the meta-learning model on a low-data regime. In addition, this also mitigates the issue of class imbalance learning during the meta-training phase.

\subsection{Implementation Details}

\label{sec:impl}
\textbf{Network Architecture}: The network for the embedding model is trained with three different backbones. The first is Conv64 --- a simple 6-layer Convolutional Neural Network (CNN) with each convolution layer comprising 64 filters with kernel size $3 \times 3$. This is followed by a batch-normalization layer, Rectified Linear Unit (ReLU) activation, and  a $2 \times 2$ max-pooling layer. 

The second and third backbones are ResNet18 and ResNet50, respectively, without the fully-connected layers at the end. During the meta-training phase, the feature outputs from the backbones are connected to a N-neuron fully-connected layer, where N is the number of classes in the training set Dnew. This effectively sets up a supervised classification task, as described in Section \ref{sec: methods}. During the meta-testing phase, the N-neuron fully-connected layer is removed, and the trained network outputs a feature vector. This has a cardinality of 64, 1000 and 2048 for Conv64, ResNet18 and ResNet50 backbones respectively. In each meta-test episode, the network extracts a feature vector representation for each image in the support and query sets, which in turn is used to train the downstream base learner on the support set and make predictions for the query set images.  

\textbf{Training Parameters}: The experiments presented in this paper used the Adam optimizer with an initial learning rate of $1e-{03}$. The decay rates for the first and second moments of gradient are set at 0.9 and 0.95 respectively. No weight decaying is performed. A grid search was performed to determine the optimal learning and decays rates, since these hyper-parameters have a considerable impact on the model performance. Each training batch consisted of 32 samples. The embedding models were trained for 90 epochs on the ISIC dataset, and for 40 epochs on the Derm7pt dataset. The respective embedding models were then used to perform the meta-testing experiments with various few-shot settings, training and predicting with a downstream base learner for each episode of the testing phase.

\textbf{Data Augmentation}: Data augmentation techniques are adopted when training the embedding network, to allow it to learn to project to a robust and discriminative feature space. During meta-training — that is when training the embedding network, random crop, color jittering, and random horizontal flip as in \cite{lee}. For the meta-testing stage, an N-way K-shot downstream base classifier is trained for each episode. The same set of augmentations are applied here, as part of the ablation study. We use the implementations in scikit-learn [<cite 1 p2>] for implementing the base learner(s).


\section{Results and Discussion}

\begin{table}
\begin{center}
\caption{\textbf{Comparison of performance with the prototypical network --- MetaDerm}, used for skin lesion classification. All experiments were performed for 2-way classification with K-shots as decribed in the column headers. Metrics are macro-averaged over all classes. \emph{Acc} is the accuracy metric. \emph{AuRoc} is the area under receiver-operating-curve metric.}
\label{table:proto}
\begin{tabular}{cccccccc}
\hline\noalign{\smallskip}
\multirow{2}{*}{\textbf{Model}} & \multirow{2}{*}{\textbf{Metric}} & \multicolumn{3}{c}{\textbf{ISIC-18}} & \multicolumn{3}{c}{\textbf{Derm7pt}} \\
& & 1-shot & 3-shot & 5-shot & 1-shot & 3-shot & 5-shot \\
\hline
\noalign{\smallskip}
\multirow{2}{*}{MetaDerm \cite{mahajan2020}} & Acc & 59.3 & 67.9 & 73.0 & 62.5 & 63.9 & 66.7 \\
& AuRoc & 61.6 & 70.2 & 75.4 & 60.2 & 65.7 & 70.5 \\
\hline
\multirow{2}{*}{Ours-Conv64-LR} & Acc & 58.9 & 68.3 & 73.1 & 62.3 & 64.6 & 67.0 \\
& AuRoc & 69.1 & 72.3 & 80.6 & 63.1 & 70.3 & 79.8 \\
\hline
\multirow{2}{*}{Ours-Conv64-SVM} & Acc & 59.2 & 69.5 & 73.8 & 63.1 & 66.2 & 67.4 \\
& AuRoc & 71.4 & 72.1 & 82.1 & 70.3 & 74.2 & 78.8 \\
\hline
\multirow{2}{*}{Ours-ResNet18-SVM} & Acc & 59.9 & 74.1 & 78.2 & 63.7 & 69.8 & 70.2 \\
& AuRoc & 72.3 & 76.5 & 79.8 & 72.2 & 74.1 & 81.1 \\
\hline
\multirow{2}{*}{Ours-ResNet50-SVM} & Acc & 65.9 & 76.5 & 79.6 & 64.0 & 74.3 & 78.1 \\
& AuRoc & 72.6 & 77.4 & 80.1 & 72.9 & 78.6 & 83.2 \\
\hline
\end{tabular}
\end{center}
\end{table}

Table \ref{table:proto} compares the model trained using the proposed meta-learning algorithm with an implementation of prototypical networks. It is evident that the proposed base learner approach consistently outperforms the prototypical network, even when using the same backbone network. In terms of accuracy, there is about 0.5-2\% improvement, and much more in terms of the AuRoc metric. It is worth noting that employing more complex backbones --- ResNet18 and ResNet50, shows further improvements. This suggests that the embedding network is able to find a more representative feature space with a more complex network, when trained using the proposed meta-training method. However, the Logistic Regression (LR) and Support Vector Machine (SVM) classifiers exhibit lower performance that the prototypical network in the 1-shot regime. This is due to lesser number of training examples available to the downstream base learner in these cases, which is not a concern in prototypical networks that only average the feature vectors across all samples of a class.

\begin{table}
\begin{center}
\caption{\textbf{Ablation study with Conv64 network as the backbone using accuracy metric.} \emph{DT} refers to a decision tree classifier used as the base learner. \emph{NN} describes a nearest neighbor classifier. \emph{LR} describes Logistic Regression classifier. \emph{L2-Norm} refers to normalization of the embedding vector to a unit hypersphere. \emph{Aug} refers to the application of the five augmentation techniques describes in Section \ref{sec:impl} on each image of the support set. All experiments are performed for 2-way classification.}
\label{table:ablation}
\begin{tabular}{ccccc cc cc}
\hline\noalign{\smallskip}
\multirow{2}{*}{\textbf{DT}} & \multirow{2}{*}{\textbf{NN}} & \multirow{2}{*}{\textbf{LR}} & \multirow{2}{*}{\textbf{L2-Norm}} & \multirow{2}{*}{\textbf{Aug}} & \multicolumn{2}{c}{\textbf{ISIC-18}} & \multicolumn{2}{c}{\textbf{Derm7pt}} \\
& & & & & 1-shot & 5-shot & 1-shot & 5-shot \\
\hline
\checkmark & & & & & 56.9 & 60.3 & 59.4 & 62.3 \\
\checkmark & & & & \checkmark & 57.1 & 61.1 & 60.1 & 62.3 \\
& \checkmark & & & & 60.4 & 67.8 & 62.4 & 63.1 \\
& \checkmark & & & \checkmark & 60.7 & 67.7 & 62.5 & 63.4 \\
& & \checkmark & & & 58.9 & 73.1 & 62.3 & 67.0 \\
& & \checkmark & \checkmark & & 59.1 & 73.4 & 62.5 & 67.4 \\
& & \checkmark & \checkmark & \checkmark & 62.4 & 75.2 & 64.4 & 68.9 \\
\hline
\end{tabular}
\end{center}
\end{table}

The ablation study is performed on the Conv64 backbone for the embedding network by applying combinations of one or more of the following elements of the proposed classification pipeline: (1) A Decision Tree (DT) classifier is used as the downstream base learner; (2) A Nearest Neighbor (NN) classifier is used as the base learner; (3) A Logistic Regression (LR) classifier is used as the base learner; (4) L2 normalization is applied on the embedding vector obtained from the trained backbone during the meta-test phase to capsulate the feature values into a unit hypersphere; (5) Five augmentation methods are applpied to each image in the support set to increase the size of the training data available to the base learner during their meta-testing phase. In effect, each image of the support set is used to produce 5 more images. Techniques used include random crop, color jittering, random horizontal and vertical flipping and random rotations of upto 10 degrees in either direction.

In general, it can be noted from Table \ref{table:ablation} that normalization and augmentation have reasonable impact in improving the performance of the LR classifier. The increase in performance with augmentation is higher for the 1-shot case. This is potentially due to stark increase in number of samples available for training the base learner due to augmentation. With the 5-shot case, the effect of more augmentations atop 5 samples per class is less pronounced. DT and NN classifiers, in general, perform well in the 1-shot regime, and their effect numbs down with increasing number of samples. This is intuitive, and is comparable with the characteristically lower margins between the performance of the proposed method and prototypical networks for the 1-shot case (refer to Table \ref{table:proto}). Finally, augmentations have little or no effect on the performance of NN and DT classifiers.

\section{Conclusion and Future Work}{

This paper presents a baseline learning approach for few-shot image classification in the meta-learning context. This approach has been underappreciated in the literature thus far, and this is the first application of this method for skin image classification in the few-shot regime. The proposed approach consistently outperforms the protoypical network on two different datasets, even when they adopt the same backbone network.

Furthermore, this paper posits empirically that even a simple linear model suffices to generalize well for a few-shot learning task, as long as a good representation of the data is given prepared --- the embedding network.The efficacy in the proposed meta-training approach may be traced back to the merging of all the N, K-way classification episodes of the meta-training dataset. Training on this single but harder N K-way classification task proves to learn a more generalizable and representative feature space through the embedding network. Further, the use of a single larger task for training also facilitates the effective use of more complex backbones without the concern of overfitting, and can learn more complex embedding spaces that transfer well to meta-testing set.

Our future work will pivot around methods to improve the few-shot classification performance, specifically by improving the quality of the embeddings through different backbone networks; experiment with different downstream classifiers; prospect the potential of the proposed method to other logical objectives such as image segmentation; and test out the technique on more complex medical datasets. A viable step in improving the quality of embeddings produced direction could be to distill the embedding network to refine the learnt feature space.
}

\bibliographystyle{splncs04}
\bibliography{references.bib} 

\begin{thebibliography}{10}
\providecommand{\url}[1]{\texttt{#1}}
\providecommand{\urlprefix}{URL }
\providecommand{\doi}[1]{https://doi.org/#1}

\bibitem{afham}
Afham, M., Khan, S.H., Khan, M.H., Naseer, M., Khan, F.S.: Rich semantics
  improve few-shot learning. CoRR  \textbf{abs/2104.12709} (2021),
  \url{https://arxiv.org/abs/2104.12709}

\bibitem{isic18t3}
Codella, N., Rotemberg, V., Tschandl, P., Celebi, M.E., Dusza, S., Gutman, D.,
  Helba, B., Kalloo, A., Liopyris, K., Marchetti, M., et~al.: Skin lesion
  analysis toward melanoma detection 2018: A challenge hosted by the
  international skin imaging collaboration (isic). arXiv preprint
  arXiv:1902.03368  (2019)

\bibitem{deng}
Deng, J., Dong, W., Socher, R., Li, L.J., Li, K., Fei-Fei, L.: Imagenet: A
  large-scale hierarchical image database. In: 2009 IEEE Conference on Computer
  Vision and Pattern Recognition. pp. 248--255 (2009).
  \doi{10.1109/CVPR.2009.5206848}

\bibitem{guneet}
Dhillon, G.S., Chaudhari, P., Ravichandran, A., Soatto, S.: A baseline for
  few-shot image classification. CoRR  \textbf{abs/1909.02729} (2019),
  \url{http://arxiv.org/abs/1909.02729}

\bibitem{finn}
Finn, C., Abbeel, P., Levine, S.: Model-agnostic meta-learning for fast
  adaptation of deep networks. CoRR  \textbf{abs/1703.03400} (2017),
  \url{http://arxiv.org/abs/1703.03400}

\bibitem{resnet}
He, K., Zhang, X., Ren, S., Sun, J.: Deep residual learning for image
  recognition (2015). \doi{10.48550/ARXIV.1512.03385},
  \url{https://arxiv.org/abs/1512.03385}

\bibitem{he}
He, K., Zhang, X., Ren, S., Sun, J.: Deep residual learning for image
  recognition. In: 2016 IEEE Conference on Computer Vision and Pattern
  Recognition (CVPR). pp. 770--778 (2016). \doi{10.1109/CVPR.2016.90}

\bibitem{derm7pt}
Kawahara, J., Daneshvar, S., Argenziano, G., Hamarneh, G.: Seven-point
  checklist and skin lesion classification using multitask multimodal neural
  nets. IEEE Journal of Biomedical and Health Informatics  \textbf{23}(2),
  538--546 (mar 2019). \doi{10.1109/JBHI.2018.2824327}

\bibitem{kim}
Kim, J., Kim, T., Kim, S., Yoo, C.D.: Edge-labeling graph neural network for
  few-shot learning. CoRR  \textbf{abs/1905.01436} (2019),
  \url{http://arxiv.org/abs/1905.01436}

\bibitem{Koch}
Koch, G.R.: Siamese neural networks for one-shot image recognition (2015)

\bibitem{laenen}
Laenen, S., Bertinetto, L.: On episodes, prototypical networks, and few-shot
  learning. CoRR  \textbf{abs/2012.09831} (2020),
  \url{https://arxiv.org/abs/2012.09831}

\bibitem{lee}
Lee, K., Maji, S., Ravichandran, A., Soatto, S.: Meta-learning with
  differentiable convex optimization. pp. 10649--10657 (06 2019).
  \doi{10.1109/CVPR.2019.01091}

\bibitem{mahajan2020}
Mahajan, K., Sharma, M., Vig, L.: Meta-dermdiagnosis: Few-shot skin disease
  identification using meta-learning. In: Proceedings of the IEEE/CVF
  Conference on Computer Vision and Pattern Recognition Workshops. pp. 730--731
  (2020)

\bibitem{boris}
Oreshkin, B.N., L{\'{o}}pez, P.R., Lacoste, A.: {TADAM:} task dependent
  adaptive metric for improved few-shot learning. CoRR  \textbf{abs/1805.10123}
  (2018), \url{http://arxiv.org/abs/1805.10123}

\bibitem{anirudh}
Raghu, A., Raghu, M., Bengio, S., Vinyals, O.: Rapid learning or feature reuse?
  towards understanding the effectiveness of {MAML}. CoRR
  \textbf{abs/1909.09157} (2019), \url{http://arxiv.org/abs/1909.09157}

\bibitem{ravi}
Ravi, S., Larochelle, H.: Optimization as a model for few-shot learning. In:
  ICLR (2017)

\bibitem{andrei}
Rusu, A.A., Rao, D., Sygnowski, J., Vinyals, O., Pascanu, R., Osindero, S.,
  Hadsell, R.: Meta-learning with latent embedding optimization. CoRR
  \textbf{abs/1807.05960} (2018), \url{http://arxiv.org/abs/1807.05960}

\bibitem{p2}
Snell, J., Swersky, K., Zemel, R.: Prototypical networks for few-shot learning.
  In: Guyon, I., Luxburg, U.V., Bengio, S., Wallach, H., Fergus, R.,
  Vishwanathan, S., Garnett, R. (eds.) Advances in Neural Information
  Processing Systems. vol.~30. Curran Associates, Inc. (2017),
  \url{https://proceedings.neurips.cc/paper/2017/file/cb8da6767461f2812ae4290eac7cbc42-Paper.pdf}

\bibitem{szegedy}
Szegedy, C., Vanhoucke, V., Ioffe, S., Shlens, J., Wojna, Z.: Rethinking the
  inception architecture for computer vision. In: 2016 IEEE Conference on
  Computer Vision and Pattern Recognition (CVPR). pp. 2818--2826 (2016).
  \doi{10.1109/CVPR.2016.308}

\bibitem{tian}
Tian, Y., Wang, Y., Krishnan, D., Tenenbaum, J.B., Isola, P.: Rethinking
  few-shot image classification: a good embedding is all you need? CoRR
  \textbf{abs/2003.11539} (2020), \url{https://arxiv.org/abs/2003.11539}

\bibitem{p1}
Vinyals, O., Blundell, C., Lillicrap, T., Kavukcuoglu, K., Wierstra, D.:
  Matching networks for one shot learning. In: Proceedings of the 30th
  International Conference on Neural Information Processing Systems. p.
  3637–3645. NIPS'16, Curran Associates Inc., Red Hook, NY, USA (2016)

\bibitem{oriol}
Vinyals, O., Blundell, C., Lillicrap, T.P., Kavukcuoglu, K., Wierstra, D.:
  Matching networks for one shot learning. CoRR  \textbf{abs/1606.04080}
  (2016), \url{http://arxiv.org/abs/1606.04080}

\bibitem{Zhou}
Zhou, B., Lapedriza, A., Khosla, A., Oliva, A., Torralba, A.: Places: A 10
  million image database for scene recognition. IEEE Transactions on Pattern
  Analysis and Machine Intelligence  \textbf{40}(6),  1452--1464 (2018).
  \doi{10.1109/TPAMI.2017.2723009}

\end{thebibliography}
\end{document}